\documentclass{article} 
\usepackage[preprint]{colm2026_conference} 

\usepackage[T1]{fontenc}
\usepackage{microtype}
\usepackage{hyperref}
\usepackage{url}
\usepackage{booktabs}

\usepackage{lineno}

\definecolor{lavender}{RGB}{200, 190, 230}
\definecolor{sandboxcolor}{HTML}{EAE8FD}
\definecolor{sandboxpink}{RGB}{255, 220, 230}
\definecolor{lavender_text}{RGB}{80, 20, 180}  

\definecolor{morandigray}{RGB}{200, 200, 195}
\definecolor{morandiblue}{RGB}{180, 200, 220}
\definecolor{morandigreen}{RGB}{170, 195, 175}
\definecolor{morandipink}{RGB}{210, 180, 190}

\newcommand{\dq}{{\usefont{T1}{cmtt}{m}{n}\symbol{34}}}

\definecolor{darkblue}{rgb}{0, 0, 0.5}
\hypersetup{colorlinks=true, citecolor=darkblue, linkcolor=darkblue, urlcolor=darkblue}

\usepackage{enumitem}
\setlist[itemize]{noitemsep, topsep=0pt}

\usepackage{tcolorbox}
\usepackage{xcolor}
\usepackage{colortbl}
\usepackage{multirow}
\usepackage{booktabs}
\usepackage{graphicx}
\usepackage{wrapfig}
\usepackage{makecell}

\usepackage{textgreek}
\usepackage{amsmath}
\usepackage{amssymb}
\usepackage{comment}
\usepackage{tikz}
\usepackage{algorithm}
\usepackage{algorithmic}
\usetikzlibrary{shapes.geometric, arrows.meta, positioning, fit, backgrounds, calc}

\usepackage{pifont} 

\makeatletter
\def\thanks#1{\protected@xdef\@thanks{\@thanks
        \protect\footnotetext{#1}}}
\makeatother

\newcommand\ours{LLM-in-Sandbox}
\newcommand\oursrl{LLM-in-Sandbox-RL}
\newcommand\oursbox{\tcbox[on line, colback=sandboxcolor, colframe=lavender_text, boxrule=0.8pt, arc=2pt, boxsep=1pt, left=1pt, right=1pt, top=0pt, bottom=0pt]{LLM}}

\title{Computer Environments Elicit\\General Agentic Intelligence in LLMs}

\author{\textbf{Daixuan Cheng}\textsuperscript{\textalpha}\textsuperscript{\textbeta}\thanks{Email: \texttt{daixuancheng6@gmail.com}~~\textsuperscript{\dag}Corresponding Authors.}~~\textbf{Shaohan Huang}\textsuperscript{\textbeta}~~\textbf{Yuxian Gu}\textsuperscript{\textgamma}~~\textbf{Huatong Song}\textsuperscript{\textalpha}~~\textbf{Guoxin Chen}\textsuperscript{\textalpha}\\[3pt]
\textbf{Li Dong}\textsuperscript{\textbeta}~~\textbf{Wayne Xin Zhao}\textsuperscript{\textalpha}\textsuperscript{\dag}~~\textbf{Ji-Rong Wen}\textsuperscript{\textalpha}~~\textbf{Furu Wei}\textsuperscript{\textbeta}\textsuperscript{\dag}
\\[6pt]
\textsuperscript{\textalpha}GSAI, Renmin University of China\quad\textsuperscript{\textbeta}Microsoft Research\quad\textsuperscript{\textgamma}Tsinghua University
\\[3pt]
~\textcolor{lavender_text}{https://llm-in-sandbox.github.io}
\\[-15pt]
}

\begin{document}

\ifcolmsubmission
\linenumbers
\fi

\maketitle

\begin{abstract}
Agentic intelligence in large language models (LLMs) requires not only model intrinsic capabilities but also interactions with external environments. Equipping LLMs with computers now represents a prevailing trend. However, the computer environment's intrinsic value has not been systematically investigated, particularly its potential to elicit general capabilities. Here we introduce \ours, which virtualizes the computer as a code sandbox with only basic functionalities, and demonstrate that this minimal setting elicits computer-based meta-capabilities for general task solving: external resource access, file management, and code execution. Without additional training, strong models achieve substantial gains (up to 15.5\%) across mathematics, physics, chemistry, biomedicine, long-context understanding, and instruction following, while reducing token consumption by up to 8 times. Furthermore, we develop \oursrl~to train models exclusively on non-agentic data within the sandbox, empowering weaker models to harness the environment and internalize these interactions. Our results demonstrate that computer environments elicit general intelligence, yield efficiency gains, and can be harnessed through training, serving as a promising foundation for generalist agents.
\end{abstract} 

\begin{figure}[!htb]
\vspace{-15pt}
\begin{center}
\includegraphics[width=0.95\textwidth]{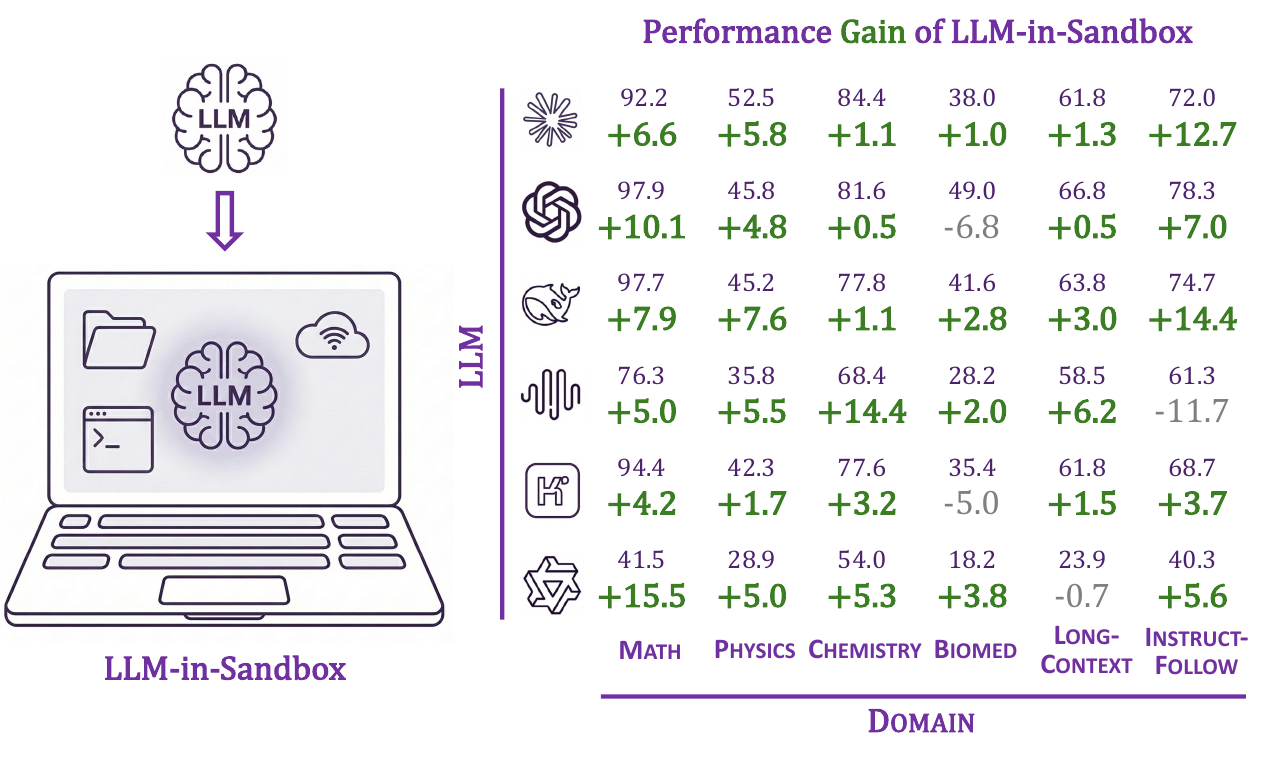}
\end{center}
\vspace{-15pt}
\caption{Overview of \ours. We enable LLMs to explore within a code sandbox (i.e., virtual computer), unlocking significant performance gains across diverse LLMs and domains. Green values indicate improvements over vanilla LLMs. \textbf{All LLMs are evaluated without additional training.}}
\label{fig:overview}
\end{figure}

\section{Introduction}
\vspace{-10pt}
The evolution toward general agents is a driving force in artificial intelligence. Large language models (LLMs) have demonstrated increasingly versatile capabilities~\citep{gpt3,o1,r1,claude_sonnet}, making them a promising foundation for this pursuit. However, the intrinsic capability of the model alone does not fully constitute an agent. Intelligence is fundamentally situated; it is shaped by how a model engages with an external environment to retrieve information, take actions, and adapt to feedback~\citep{wiener1949cybernetics,russell1995modern,sutton1998reinforcement}. From this perspective, interacting with the environment acts as a defining architectural component: it dictates whether the model's intrinsic power can be translated into effective task-solving across diverse domains.

In the real world, computers serve as a primary platform for such interaction, providing a universal interface for humans to execute complex intellectual tasks. This versatility makes computer environments a promising foundation for grounding LLM agents. Indeed, a wave of agentic systems such as Manus~\citep{manus}, Claude Code~\citep{claude_code} and OpenClaw~\citep{openclaw} have begun equipping LLMs with computer environments, reflecting surging interest in this direction. However, these systems interleave basic environment interaction with high-level interfaces, pre-built capabilities, and other system-level optimizations, making the source of their performance difficult to isolate. Whether the observed gains arise primarily from the environment itself, or from the broader system design built around it, is not well understood. Moreover, whether models can learn to leverage these environments, and thereby acquire genuinely generalizable agentic intelligence, has yet to be systematically examined.

Here we introduce \ours~(Figure~\ref{fig:overview}) to investigate whether the computer, virtualized as a code sandbox, can elicit general agentic intelligence in LLMs. Compared to the standard inference setting where the model directly produces output without environmental interaction, which we refer to as LLM mode, \ours~mode allows the model to interact with the sandbox through multi-turn tool calls. A code sandbox is an isolated virtual environment for terminal access, file management, and code execution.
To disentangle the role of the environment from other confounding factors, our design follows two principles: \textit{minimal}, preserving only basic computer functionalities without any domain-specific additions; and \textit{exploratory}, encouraging models to freely experiment within this environment. Code sandboxes have proven effective for coding tasks~\citep{sweagent}; here we focus on non-code domains to test whether their benefits extend beyond coding. We further propose \oursrl, a reinforcement learning method that trains models within the sandbox using only non-agentic data, to investigate whether models can leverage such environments to acquire more general capabilities through learning.

We evaluate \ours~inference mode using seven frontier models across six challenging domains, spanning mathematics, physics, chemistry, biomedicine, long-context understanding, and instruction following. Remarkably, without any additional training, strong models achieve consistent performance gains across all evaluated domains, with improvements reaching +15.5\% in mathematics and +14.4\% in instruction following. Trajectory distributions reveal that these gains manifest through three environment-elicited meta-capabilities: external resource access, file management, and code execution. For example, models autonomously fetch external resources to acquire domain-specific knowledge, manage files to process documents exceeding their context windows, and execute code to meticulously verify output correctness. Crucially, these capabilities emerge naturally from model-environment interaction rather than pre-defined workflows, providing direct evidence that computer environments inherently elicit new agentic behaviors. Furthermore, this environmental grounding sharply improves efficiency: by offloading extensive information to the file system of the computer environment, token consumption falls by up to 8$\times$ in long-context scenarios with minimal infrastructure overhead. Ultimately, this paradigm delivers enhanced capability at reduced cost, firmly establishing computer environments as a universally potent foundation for LLMs.

While strong models benefit directly from the computer environment, weaker models often struggle, performing worse in \ours~mode than in LLM mode. To examine whether training can bridge this gap, we propose \oursrl, which applies reinforcement learning within the computer environment. The training data uses only general, non-agentic context-based tasks~\citep{instructpt} that do not target coding or any evaluation benchmark. Furthermore, background materials (i.e., the context) are stored as files in the environment rather than provided in the model prompt, forcing models to actively explore the environment to complete each task. Together with the minimal sandbox and the absence of any downstream-specific optimization, this design ensures that any performance gains can be more credibly attributed to the model learning to interact with the environment. Experiments show that, with only outcome-based rewards~\citep{r1}, even this simplified training setup generalizes across all evaluated domains, spanning both non-code tasks and software engineering; weaker models, in particular, shift from underperforming LLM mode to substantially outperforming it in \ours~mode, while strong models also gain further improvements. Moreover, we find that \oursrl~also improves the model itself, even in LLM mode without environmental access, because structured decomposition and self-verification developed through environmental interaction are internalized into its text-only reasoning.

Overall, our results show the computer environment provides a systematic mechanism to elicit broader capabilities in LLMs. We observe gains in both training-free and post-trained settings across task domains and model capability levels, with post-training gains extending even to settings without environmental access. These gains also come with substantial efficiency advantages, showing that capability and efficiency can improve together. Given that these results point to computer environments as a promising foundation for next-generation LLMs, we release \ours~as an open-source library, compatible with popular inference backends (e.g., vLLM~\citep{vllm} and SGLang~\citep{sglang}) and API-based LLMs, to support future research and real-world deployment.

\section{Computer Environments Elicit General Intelligence}
\label{sec:ise}

The core idea of \ours~is to grant LLMs access to a computer environment where they can freely operate to complete user-specified tasks. Specifically, computers possess three meta-capabilities that form the foundation for general task-solving:
\begin{itemize}[leftmargin=*,itemsep=2pt]
    \item \textit{External resource access}: fetching resources from external services (e.g., the internet);
    \item \textit{File management}: reading, writing, and organizing data persistently;
    \item \textit{Code execution}: writing and executing arbitrary programs.
\end{itemize}
Just as humans leverage computers to accomplish virtually any task, we hypothesize that combining LLMs' powerful reasoning and agentic capabilities with a code sandbox may unlock their potential for general intelligence. 

To explore the full potential of this paradigm, our design of \ours~emphasizes two principles: \textbf{minimal}, providing a code sandbox with only basic computer functionalities, and \textbf{exploratory}, encouraging models to freely experiment within this environment. In the following, we describe our sandbox environment (Section~\ref{sec:sandbox}), the \ours~workflow (Section~\ref{sec:workflow}), as well as experiments (Section~\ref{sec:ise_exp}) and analysis (Section~\ref{sec:analysis}) in general domains.

\subsection{Virtualizing the Computer as a Code Sandbox}
\label{sec:sandbox}

A code sandbox is a virtualized computer environment, typically an Ubuntu-based system implemented via Docker containers, that provides LLMs with terminal access and full system capabilities. Within this environment, LLMs can execute arbitrary bash commands, create and modify files, and access network resources. The containerized nature ensures isolation from the host system, enabling safe execution of model-generated code. 

\begin{table}[h]
\centering
\small
\begin{tabular}{l|cc}
\toprule
& \textbf{SWE Agents} & \textbf{\ours} \\
\midrule
Environment Setup & Task-specific & General-purpose \\
Dependencies & Pre-configured & Runtime installation \\
Storage Scaling & Per-task images & Single shared image \\
\bottomrule
\end{tabular}
\vspace{-5pt}
\caption{Comparison of sandbox design between SWE agents and \ours. }
\label{tab:sandbox_comparison}
\end{table}
\paragraph{Lightweight General-Purpose Design.}
Code sandboxes have recently emerged as a critical infrastructure for code agents like Claude Code~\citep{claude_code}. However, existing sandbox-based systems, especially those for software engineering tasks~\citep{r2e,openhands,sweagent}, require complex, task-specific environments. Instead, we provide a lightweight and general-purpose environment equipped only with a standard Python interpreter and essential scientific computing libraries (e.g., NumPy, SciPy), and delegate domain-specific tool acquisition to the model itself. During execution, models can install or create any tools they deem necessary. Table~\ref{tab:sandbox_comparison} summarizes the key differences. This design offers two advantages: (1) \textit{Generalizability}: the same environment supports diverse tasks without manual reconfiguration, and (2) \textit{Scalability}: the uniform setup enables efficient large-scale inference and training without per-task overhead. For example, when scaling to thousands of tasks, SWE agents may require up to 6 TB of storage for task-specific images~\citep{swegym}, whereas our shared image approach maintains a constant footprint of only $\sim$1.1 GB. 

\paragraph{Minimal Toolset with Meta-Capabilities.}
Within the code sandbox, we equip the model with three fundamental tools that together realize the core capabilities of a computer: (1) \texttt{bash} for executing arbitrary terminal commands, (2) \texttt{file\_editor} for file creation, viewing, and editing, and (3) \texttt{finish} for indicating task completion.
Specifically, \texttt{bash} is a powerful \textit{meta-tool} that enables models to install packages, run programs, and even programmatically create new tools on demand, thereby bootstrapping any additional functionality beyond the provided toolset.
Detailed specifications are provided in Appendix~\ref{sec:sandbox_implementation}.

\subsection{\ours~Workflow}
\label{sec:workflow}
Our workflow builds on the ReAct framework~\citep{react}, where the model iteratively reasons and acts based on environmental feedback. As shown in Algorithm~\ref{alg:ise} ({\color{lavender_text}purple} highlights indicate sandbox-specific components), at each turn, the model generates a tool call, receives the execution result from the sandbox, and decides the next action. This multi-turn interaction continues until the model calls \texttt{finish} or reaches a maximum turn limit. To accommodate diverse scenarios in general tasks, our workflow encourages free exploration and supports flexible input/output handling.

\begin{algorithm}[htb]
\caption{\ours~Workflow}
\label{alg:ise}
\begin{algorithmic}[1]
\REQUIRE Task prompt $p$, {\color{lavender_text}Task requirements $r$ (optional), Sandbox $\mathcal{S}$}, Maximum turns $T$
\ENSURE Final output $o$
\STATE {\color{lavender_text}Configure sandbox $\mathcal{S}$ with task requirements $r$ (if any)}
\STATE $t \leftarrow 0$
\STATE {\color{lavender_text}Tools: \{\texttt{bash}, \texttt{file\_editor}, \texttt{finish}\}}
\WHILE{$t < T$}
    \STATE Model generates tool call $a_t$ based on prompt $p$ and history
    \IF{$a_t$ is \texttt{finish}}
        \STATE \textbf{break}
    \ENDIF
    \STATE {\color{lavender_text}Execute $a_t$ in $\mathcal{S}$, obtain observation $obs_t$}
    \STATE Append $(a_t, obs_t)$ to interaction history
    \STATE $t \leftarrow t + 1$
\ENDWHILE
\STATE {\color{lavender_text}Extract output $o$ from sandbox $\mathcal{S}$ (e.g., \texttt{/testbed/answer.txt})}
\RETURN $o$
\end{algorithmic}
\end{algorithm}

\paragraph{Prompting for Exploration.}
We design a system prompt that guides models to fully utilize the environment. First, it encourages models to leverage computational tools rather than performing calculations through natural language. Second, it emphasizes deriving answers through program execution instead of directly hardcoding results. Third, it informs models that the environment is a safe, isolated space where they can freely explore diverse approaches to complete tasks. The full system prompt is provided in Appendix~\ref{sec:appendix_prompt}.

\paragraph{Task Input/Output Handling.}
We leverage the computer's file system to flexibly handle diverse input/output formats. For inputs, content can be provided not only via the model prompt but also through files. For example, for long-context understanding tasks that require reading documents, we can regard the documents as task requirements and place documents in \texttt{/testbed/documents/}. For outputs, the model is instructed to place the final result at a designated location (e.g., \texttt{/testbed/answer.txt}), containing only the final result without intermediate content. After task completion, the result is extracted from this location as the final output. This approach cleanly separates exploration from final output and naturally accommodates various data formats. 

\begin{table}[t]
\begin{center}
\small
\setlength{\tabcolsep}{4pt}
\resizebox{\textwidth}{!}{
\begin{tabular}{l|ccc|ccc|ccc}
\toprule
\multirow{2}{*}{\textbf{Model}} & \multicolumn{3}{c|}{\textbf{Mathematics}} & \multicolumn{3}{c|}{\textbf{Physics}} & \multicolumn{3}{c}{\textbf{Chemistry}} \\
& LLM & \oursbox & $\Delta$ & LLM & \oursbox & $\Delta$ & LLM & \oursbox & $\Delta$ \\
\midrule
Claude-Sonnet-4.5-Think & 85.6 & \textbf{92.2} & \small{\textcolor{green!50!black}{+6.6}} & 46.7 & \textbf{52.5} & \small{\textcolor{green!50!black}{+5.8}} & 83.3 & \textbf{84.4} & \small{\textcolor{green!50!black}{+1.1}} \\
GPT-5 & 87.8 & \textbf{97.9} & \small{\textcolor{green!50!black}{+10.1}} & 41.0 & \textbf{45.8} & \small{\textcolor{green!50!black}{+4.8}} & 81.1 & \textbf{81.6} & \small{\textcolor{green!50!black}{+0.5}} \\
DeepSeek-V3.2-Thinking & 89.8 & \textbf{97.7} & \small{\textcolor{green!50!black}{+7.9}} & 37.6 & \textbf{45.2} & \small{\textcolor{green!50!black}{+7.6}} & 76.7 & \textbf{77.8} & \small{\textcolor{green!50!black}{+1.1}} \\
MiniMax-M2 & 71.3 & \textbf{76.3} & \small{\textcolor{green!50!black}{+5.0}} & 30.3 & \textbf{35.8} & \small{\textcolor{green!50!black}{+5.5}} & 54.0 & \textbf{68.4} & \small{\textcolor{green!50!black}{+14.4}} \\
Kimi-K2-Thinking & 90.2 & \textbf{94.4} & \small{\textcolor{green!50!black}{+4.2}} & 40.6 & \textbf{42.3} & \small{\textcolor{green!50!black}{+1.7}} & 74.4 & \textbf{77.6} & \small{\textcolor{green!50!black}{+3.2}} \\
Qwen3-Coder-30B-A3B & 26.0 & \textbf{41.5} & \small{\textcolor{green!50!black}{+15.5}} & 23.9 & \textbf{28.9} & \small{\textcolor{green!50!black}{+5.0}} & 48.7 & \textbf{54.0} & \small{\textcolor{green!50!black}{+5.3}} \\
\textcolor{black!50}{Qwen3-4B-Instruct-2507} & \textcolor{black!50}{\textbf{46.0}} & \textcolor{black!50}{32.5} & \small{\textcolor{red!50}{-13.5}} & \textcolor{black!50}{22.9} & \textcolor{black!50}{\textbf{25.6}} & \small{\textcolor{green!50!black!50}{+2.7}} & \textcolor{black!50}{\textbf{56.4}} & \textcolor{black!50}{49.3} & \small{\textcolor{red!50}{-7.1}} \\
\bottomrule
\end{tabular}
}

\vspace{3mm}

\resizebox{\textwidth}{!}{
\begin{tabular}{l|ccc|ccc|ccc}
\toprule
\multirow{2}{*}{\textbf{Model}} & \multicolumn{3}{c|}{\textbf{Biomedicine}} & \multicolumn{3}{c|}{\textbf{Long-Context}} & \multicolumn{3}{c}{\textbf{Instruct. Follow.}} \\
& LLM & \oursbox & $\Delta$ & LLM & \oursbox & $\Delta$ & LLM & \oursbox & $\Delta$ \\
\midrule
Claude-Sonnet-4.5-Think & 37.0 & \textbf{38.0} & \small{\textcolor{green!50!black}{+1.0}} & 60.5 & \textbf{61.8} & \small{\textcolor{green!50!black}{+1.3}} & 59.3 & \textbf{72.0} & \small{\textcolor{green!50!black}{+12.7}} \\
GPT-5 & \textbf{55.8} & 49.0 & \small{\textcolor{red}{-6.8}} & 66.3 & \textbf{66.8} & \small{\textcolor{green!50!black}{+0.5}} & 71.3 & \textbf{78.3} & \small{\textcolor{green!50!black}{+7.0}} \\
DeepSeek-V3.2-Thinking & 38.8 & \textbf{41.6} & \small{\textcolor{green!50!black}{+2.8}} & 60.8 & \textbf{63.8} & \small{\textcolor{green!50!black}{+3.0}} & 60.3 & \textbf{74.7} & \small{\textcolor{green!50!black}{+14.4}} \\
MiniMax-M2 & 26.2 & \textbf{28.2} & \small{\textcolor{green!50!black}{+2.0}} & 52.3 & \textbf{58.5} & \small{\textcolor{green!50!black}{+6.2}} & \textbf{73.0} & 61.3 & \small{\textcolor{red}{-11.7}} \\
Kimi-K2-Thinking & \textbf{40.4} & 35.4 & \small{\textcolor{red}{-5.0}} & 60.3 & \textbf{61.8} & \small{\textcolor{green!50!black}{+1.5}} & 65.0 & \textbf{68.7} & \small{\textcolor{green!50!black}{+3.7}} \\
Qwen3-Coder-30B-A3B & 14.4 & \textbf{18.2} & \small{\textcolor{green!50!black}{+3.8}} & 24.6 & \textbf{23.9} & \small{\textcolor{red}{-0.7}} & 34.7 & \textbf{40.3} & \small{\textcolor{green!50!black}{+5.6}} \\
\textcolor{black!50}{Qwen3-4B-Instruct-2507} & \textcolor{black!50}{10.2} & \textcolor{black!50}{\textbf{9.4}} & \small{\textcolor{red!50}{-0.8}} & \textcolor{black!50}{8.6} & \textcolor{black!50}{\textbf{10.5}} & \small{\textcolor{green!50!black!50}{+1.9}} & \textcolor{black!50}{\textbf{33.7}} & \textcolor{black!50}{29.0} & \small{\textcolor{red!50}{-4.7}} \\
\bottomrule
\end{tabular}
}
\vspace{-5pt}
\caption{Task performance of models under LLM and \ours~generation modes across domains. \oursbox~denotes \ours~mode. $\Delta$ = \ours~$-$ LLM denotes the performance difference of \ours~relative to LLM.}
\label{tab:ise_results}
\end{center}
\end{table}

\subsection{General Task Performance in Computer Environments}
\label{sec:ise_exp}
We conduct experiments to investigate whether computer environment access improves LLM performance on general tasks. Below we present the experimental setup and results.

\paragraph{Setup}
We compare \ours~with vanilla LLM generation (i.e., directly generating the output without environmental access) across diverse models and domains. The evaluated LLMs cover frontier proprietary, open-weight, code-specialized, and smaller general-purpose models: Claude-Sonnet-4.5-Thinking~\citep{claude_sonnet}, GPT-5~\citep{gpt5}, DeepSeek-V3.2-Thinking~\citep{deepseekv3.2}, MiniMax-M2~\citep{minimaxm2}, Kimi-K2-Thinking~\citep{kimik2}, Qwen3-Coder-30B-A3B-Instruct~\citep{qwen3}, and Qwen3-4B-Instruct-2507~\citep{qwen3}. 

We test on challenging tasks in six domains: Mathematics, Physics, Chemistry, Biomedicine, Long-Context Understanding, and Instruction Following. For long-context tasks, we store the input documents in the sandbox rather than including them in the prompt, to test the model's ability to leverage the environment. Since models have internet access in the sandbox, we reframe test problems to prevent benchmark hacking and manually verify sampled trajectories to ensure valid reasoning. The detailed environment implementations, model configurations, and evaluation protocols are in Appendices~\ref{sec:sandbox_implementation}-\ref{app:evaluation_details}.

\paragraph{Results}
As shown in Table~\ref{tab:ise_results}, strong agentic models consistently benefit from \ours, with improvements observed across all evaluated domains: from computation-intensive tasks (Mathematics) to knowledge-intensive tasks (Chemistry, Biomedicine) to general capabilities (Instruction Following, Long-Context). The largest gains reach +15.5\% (Qwen3-Coder on Mathematics). However, weaker models like Qwen3-4B-Instruct fail to benefit and even perform worse. We analyze the reasons in the following sections.

\subsection{Computer Capability Usage Across Domains and Model Strengths}
\label{sec:analysis}
To understand how models leverage the computer environment, we conduct a case study and quantitative analysis. Specifically, we find three computer-based meta-capabilities that benefit general task-solving: \textit{external resource access}, \textit{file management}, and \textit{code execution}. Code execution can serve many purposes; here we specifically track \textit{computation}-oriented operations. We identify these behaviors through pattern matching on model actions: (1) \textit{external resources}: network requests (e.g., \texttt{curl}, \texttt{requests.get}) and package installation (e.g., \texttt{pip install}); (2) \textit{file management}: file I/O operations (e.g., \texttt{open()}, \texttt{json.load}) and shell commands (e.g., \texttt{cat}, \texttt{grep}); (3) \textit{computation}: numerical solvers, iterative algorithms, and simulation loops. Detailed classification patterns are provided in Appendix~\ref{app:capability_classification}.

\subsubsection{Case Study}
\label{sec:case_study}
We conduct a case study to illustrate how strong agentic models utilize the computer's core capabilities to solve non-code tasks, where strong models refer to those that perform better in \ours~mode than in LLM mode in Table\ref{tab:ise_results}. Below, we present representative trajectory snippets and \textcolor{gray}{\textit{Gray text indicates our commentary}}.

\paragraph{External Resources Access.}
In the chemistry task, the model is asked to predict molecular properties given only compound names. To solve this, the model autonomously installs Java runtime via \texttt{apt-get} and downloads the OPSIN library to convert chemical names to molecular structures, acquiring domain-specific tools unavailable in the base environment.

\begin{tcolorbox}[
    title={Chemistry Trajectory (External Resources Access)},
    colback=white,
    colframe=lavender,
    coltitle=gray!40!black,
    fonttitle=\bfseries,
    arc=1mm,
    boxrule=0.6mm,
    left=1mm,
    right=1mm,
    top=1mm,
    bottom=1mm,
]
\small
\ttfamily
\textbf{[Turn 1] Action:} bash `pip install rdkit-pypi -q'\\
\textcolor{gray}{\textnormal{\textit{// RDKit installed but fails to import due to NumPy version conflicts}}}\\
\\
\textbf{[Turn 8] Action:} bash `apt-get install -y default-jre'\\
\textbf{[Turn 8] Observation:} Setting up openjdk-21-jre-headless .\kern-0.1em.\kern-0.1em. done\\
\\
\textbf{[Turn 12] Action:} bash `curl -s \textbackslash\\
\hspace*{2em} https://api.github.com/repos/dan2097/opsin/releases/latest | grep tag\_name'\\
\textbf{[Turn 12] Observation:} \dq tag\_name\dq: \dq 2.8.0\dq\\
\\
\textbf{[Turn 14] Action:} bash `apt-get install -y libopsin-java'\\
\textbf{[Turn 14] Observation:} Setting up libopsin-java (2.8.0-3) .\kern-0.1em.\kern-0.1em. done\\
\\
\textbf{[Turn 18] Action:} bash `java -jar opsin-cli.jar -o smi \textless{} input.txt'\\
\textbf{[Turn 18] Observation:} NCC(CCCCCN1/C(/C(C2=CC=CC=C12).\kern-0.1em.\kern-0.1em.  \textcolor{gray}{\textnormal{\textit{// molecular structure}}}
\end{tcolorbox}

\paragraph{File Management.}
In the long-context task, the model is given lengthy industry reports (exceeding 100K tokens) and asked to extract specific information. Rather than processing entire documents in memory, the model uses shell tools like \texttt{grep} and \texttt{sed} to locate relevant sections, then writes Python scripts to systematically extract the information.

\begin{tcolorbox}[
    title={Long-Context Understanding Trajectory (File Management)},
    colback=white,
    colframe=morandigray,
    coltitle=gray!40!black,
    fonttitle=\bfseries,
    arc=1mm,
    boxrule=0.6mm,
    left=1mm,
    right=1mm,
    top=1mm,
    bottom=1mm,
]
\small
\ttfamily
\textbf{[Turn 5] Action:} bash `ls -la /testbed/documents/'\\
\textbf{Observation:}\\
\textnormal{DAF-COMP(2023)14.en.txt \hspace{1em} 146781 bytes}\\
\textnormal{The-State-of-Competition.txt \hspace{1em} 29048 bytes .\kern0.15em.\kern0.15em.}\\[-3pt]
\\
\textbf{[Turn 8] Action:} bash `grep -n -i `infringement' .\kern-0.1em.\kern-0.1em.'\\
\textbf{Observation:}\\
\textnormal{26: .\kern0.15em.\kern0.15em. as measured by infringement notices issued by the .\kern0.15em.\kern0.15em.}\\
\textnormal{252: .\kern0.15em.\kern0.15em. estimate the number of consumer-related infringements .\kern0.15em.\kern0.15em.}\\
\\
\textbf{[Turn 9] Action:} bash `sed -n `240,280p' .\kern-0.1em.\kern-0.1em.'\\
\textcolor{gray}{\textnormal{\textit{// Jump to line range based on grep results}}}\\
\textbf{Observation:} \textnormal{There have been 12 infringement notices .\kern0.15em.\kern0.15em.}\\
\\
\textbf{[Turn 15] Action:} file\_editor create extract\_industries.py\\
\textcolor{gray}{\textnormal{\textit{// Write script to extract industry-infringement pairs}}}\\
\textnormal{def find\_sentences\_with\_infringement\_and\_industry(text):}\\
\textnormal{\hspace{1em}sentences = re.split(r`[.!?]', text)}\\
\textnormal{\hspace{1em}.\kern0.15em.\kern0.15em.}\\
\textnormal{industry\_matches = re.findall(r`(\textbackslash b[\textbackslash w\textbackslash s]+)industry', .\kern0.15em.\kern0.15em.)}
\end{tcolorbox}

\paragraph{Code Execution.}
In the instruction-following task, the model is asked to generate three sentences about medieval history, with a strict constraint: all sentences must have the \textit{same character count} while using \textit{entirely different words}. To satisfy this constraint, the model writes Python scripts to count characters, detect word overlaps, and iteratively refine candidate sentences, a task that would be extremely difficult through pure text generation.

\begin{tcolorbox}[
    title={Instruction Following Trajectory (Code Execution)},
    colback=white,
    colframe=morandiblue,
    coltitle=gray!40!black,
    fonttitle=\bfseries,
    arc=1mm,
    boxrule=0.6mm,
    left=1mm,
    right=1mm,
    top=1mm,
    bottom=1mm,
]
\small
\ttfamily
\textbf{[Turn 3] Action:} file\_editor create helper.py\\
\textcolor{gray}{\textnormal{\textit{// Count characters and extract word sets}}}\\
\textnormal{def char\_count(s): return len(s)}\\
\textnormal{def word\_set(s): return set(re.findall(r'\textbackslash b\textbackslash w+\textbackslash b', s.lower()))}\\
\\
\textbf{[Turn 4] Action:} bash `python3 helper.py'\\
\textbf{Observation:}\\
\textnormal{Lengths: [90, 95, 91]}\\
\textnormal{Overlap between 0 and 1: \{`the'\}}\\
\textcolor{gray}{\textnormal{\textit{// Constraints not satisfied}}}\\
\\
\textbf{[Turn 12] Action:} file\_editor create generate.py\\
\textcolor{gray}{\textnormal{\textit{// Combinatorial search over sentence templates}}}\\
\textnormal{for s1, s2, s3 in itertools.product(political, religious, legal):}\\
\textnormal{\hspace{1em}if check\_constraints([s1, s2, s3]): candidates.append(...)}\\
\\
\textbf{[Turn 13] Action:} bash `python3 generate.py'\\
\textbf{Observation:}\\
\textnormal{Found 363 candidates with equal length and no overlapping words}
\end{tcolorbox}

\subsubsection{Quantitative Analysis}
\label{sec:ise_analysis}
We quantify these behavioral patterns to analyze how computer utilization varies across different models and task domains. Using the pattern matching approach described above, we measure how frequently models invoke each of the three core capabilities during their interactions, computed as the ratio of capability invocations to total interaction turns. 

\begin{figure}[!htb]
\begin{center}
\includegraphics[width=\textwidth]{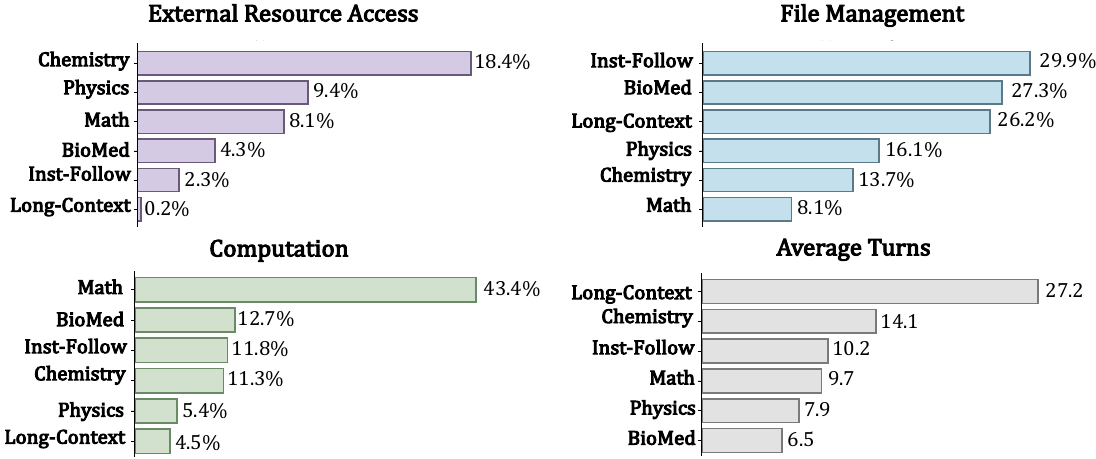}
\end{center}
\vspace{-5pt}
\caption{Computer capability behavior patterns across task domains for strong agentic models. (a)-(c): Capability usage rate, computed as capability invocations / total turns. (d): Average number of interaction turns per task.}
\label{fig:domain_analysis}
\end{figure}

\paragraph{Variation across Task Domains.}
Figure~\ref{fig:domain_analysis} shows strong models adapt their usage patterns to task requirements. \textit{Mathematics} benefits most due to high computation frequency (43.4\%): models verify solutions through numerical computation. \textit{Chemistry} shows the highest external resource frequency (18.4\%) as models install domain-specific packages. 
In contrast, \textit{Biomedicine} shows less stable improvements with the shortest exploration (6.5 turns), suggesting that models fail to fully leverage the computer environment for these tasks.

\begin{wraptable}{r}{0.4\textwidth}
\centering
\small
\begin{tabular}{lcc}
\toprule
\textbf{Model} & \textbf{Prompt} & \textbf{Env.} \\
\midrule
Claude & 11.9 & 61.8 \\
GPT-5 & 66.3 & 66.8 \\
DeepSeek & 16.8 & 63.8 \\
MiniMax & 61.0 & 58.5 \\
Kimi & 51.0 & 61.8 \\
Qwen-Coder & 30.5 & 23.9 \\
Qwen-4B & 11.8 & 10.5 \\
\midrule
\textbf{Average} & 35.6 & \textbf{49.6} \\
\bottomrule
\end{tabular}
\vspace{-5pt}
\caption{Long-context performance, both use \ours~mode but place the context differently: in prompt vs. in environment.}
\label{tab:long_context_comparison}
\vspace{-0.5em}
\end{wraptable}

\paragraph{Long-Context Tasks Benefit from File-based Context.} In Figure~\ref{fig:domain_analysis}, \textit{Long-Context} tasks show high file operation frequency with minimal external resource usage, indicating that models focus on understanding local context. To further validate the benefit of file-based context handling, we compare two settings under \ours~mode: placing documents directly in the prompt vs. storing them in the sandbox environment. As shown in Table~\ref{tab:long_context_comparison}, storing documents in the environment yields substantial gains on average, with Claude, DeepSeek, and Kimi showing the largest improvements. This suggests that \ours~serves as a promising solution for handling long-context tasks by offloading extensive data to the environment. However, performance varies across models: Qwen perform worse with environment-based context, highlighting the need for training models to effectively explore file-based information.

\paragraph{Strong vs. Weak Models.}
Table~\ref{tab:strong_weak_comparison} compares computer utilization between strong and weak models. Strong models effectively leverage all three capabilities with high usage rate (6--21\%), while the weak model (Qwen3-4B-Instruct) achieves far lower frequency ($<$3\%) despite taking nearly twice as many turns (23.7 vs. 12.6). This indicates that the weak model ``wanders'' in the environment without effective tool utilization, consuming more turns while accomplishing less.

\begin{table}[h]
\centering
\small
\begin{tabular}{l|ccc|c}
\toprule
\textbf{Model Type} & \textbf{External} & \textbf{File} & \textbf{Computation} & \textbf{Avg. Turns} \\
\midrule
Strong Models & 6.2\% & 21.1\% & 12.5\% & 12.6 \\
Weak Model & 0.8\% & 2.9\% & 2.9\% & 23.7 \\
\bottomrule
\end{tabular}
\vspace{-5pt}
\caption{Computer capability usage rate comparison: strong models (average of all models except Qwen3-4B-Instruct) vs. weak model (Qwen3-4B-Instruct). Usage rate = capability invocations / total turns.}
\label{tab:strong_weak_comparison}
\end{table}

\section{Computer Environments Enable Efficient Deployment}
\label{sec:efficiency}
We analyze practical considerations of deploying \ours~in real-world systems from two perspectives: computational analysis (Section~\ref{sec:computational_analysis}) and sandbox infrastructure overhead (Section~\ref{sec:sandbox_overhead}). We conduct experiments with local model serving across different LLMs and serving engines: DeepSeek-V3.2-Thinking and Kimi-K2-Thinking are served using SGLang~\citep{sglang}, while MiniMax-M2 and Qwen3-Coder-30B-A3B are served using vLLM~\citep{vllm}. We use a single NVIDIA DGX node for all experiments, with query concurrency set to 64, sampling the same number of task instances from each benchmark, and other settings following Section~\ref{sec:ise_exp}.

\subsection{Computational Analysis}
\label{sec:computational_analysis}

\paragraph{Cost.}
We first measure the total token consumption per query, which directly reflects the compute budget since inference FLOPs scale linearly with token count. As shown in Table~\ref{tab:query_tokens_new}, the results vary across task types. For most tasks, \ours~consumes more tokens due to multi-turn exploration. However, for long-context tasks, \ours~dramatically reduces tokens by storing content in local files rather than in the prompt. The reduction reaches up to 8$\times$ for Qwen (100K $\rightarrow$ 13K tokens). When aggregating across all tasks, \ours~consumes only \textbf{0.5--0.8$\times$} the total tokens of LLM mode.

\begin{table}[!htb]
\centering
\setlength{\tabcolsep}{4pt}
\small
\resizebox{\textwidth}{!}{
\begin{tabular}{lcccccccccccc}
\toprule
\multirow{2}{*}{\textbf{Task}} & \multicolumn{3}{c}{\textbf{DeepSeek}} & \multicolumn{3}{c}{\textbf{MiniMax}} & \multicolumn{3}{c}{\textbf{Kimi}} & \multicolumn{3}{c}{\textbf{Qwen}} \\ \cmidrule(r){2-4} \cmidrule(r){5-7} \cmidrule(r){8-10} \cmidrule(r){11-13}
 & LLM & \oursbox~& $\Delta$ & LLM & \oursbox~& $\Delta$ & LLM & \oursbox~& $\Delta$ & LLM & \oursbox~& $\Delta$ \\
\midrule
Math & 14.9 & 16.8 & +1.9 & 22.4 & 20.9 & -1.5 & 25.7 & 14.9 & -10.8 & 2.5 & 10.4 & +7.9 \\
Phy. & 8.2 & 11.6 & +3.4 & 8.1 & 7.7 & -0.4 & 10.0 & 13.6 & +3.6 & 1.5 & 6.5 & +5.0 \\
Chem. & 3.6 & 22.0 & +18.4 & 9.4 & 10.9 & +1.5 & 5.3 & 13.2 & +7.9 & 0.6 & 10.5 & +9.9 \\
Biomed. & 2.5 & 11.5 & +9.0 & 3.2 & 6.9 & +3.7 & 4.2 & 9.8 & +5.6 & 0.7 & 4.2 & +3.5 \\
Long. & 90.3 & 25.4 & \textbf{-64.9} & 88.4 & 13.6 & \textbf{-74.8} & 91.8 & 21.7 & \textbf{-70.1} & 102.9 & 12.9 & \textbf{-90.0} \\
Inst. & 2.4 & 14.9 & +12.5 & 4.2 & 8.8 & +4.6 & 6.0 & 9.6 & +3.6 & 1.6 & 8.9 & +7.3 \\
\midrule
\textbf{Average} & 20.3 & 17.0 & & 22.6 & 11.5 & & 23.8 & 13.8 & & 18.3 & 8.9 & \\
\textbf{Ratio} & \multicolumn{3}{c}{\textbf{0.84}$\times$} & \multicolumn{3}{c}{\textbf{0.51}$\times$} & \multicolumn{3}{c}{\textbf{0.58}$\times$} & \multicolumn{3}{c}{\textbf{0.49}$\times$} \\
\bottomrule
\end{tabular}
}
\vspace{-5pt}
\caption{Token consumption per query (in thousands). Each cell shows total tokens (prompt + model-generated + environment-generated tokens). $\Delta$ = \ours~$-$ LLM. The ``Ratio'' row shows $\sum N_{\text{\ours}} / \sum N_{\text{LLM}}$ computed over all tasks, reflecting the overall token savings. \oursbox~denotes \ours~mode.}
\label{tab:query_tokens_new}
\end{table}

\paragraph{Speed.}
In \ours~mode, a significant portion of tokens come from the environment, such as code execution results. Unlike model-generated tokens that require slow autoregressive decoding, environment tokens are processed via the fast \textit{Prefill}~\citep{flashattention}. As shown in Table~\ref{tab:efficiency_summary}, environment tokens constitute 37\%--51\% of the trajectory, yet environment execution accounts for less than 4\% of total time. We measure end-to-end query throughput using QPM (Queries Per Minute), i.e., the number of queries processed per unit time from submission to final answer. Overall, \ours~achieves competitive throughput: MiniMax achieves 2.2$\times$ speedup, while others range from 0.6$\times$ to 1.1$\times$.

\begin{table}[h]
\centering
\small
\begin{tabular}{lccc}
\toprule
\textbf{} & $N_{\text{env}}/N_{\text{total}}$ & $T_{\text{exe}}/T_{\text{total}}$ & {QPM Ratio} \\
\midrule
DeepSeek & 43.6\% & 2.3\% & 0.6$\times$ \\
MiniMax & 51.1\% & 2.2\% & 2.2$\times$ \\
Kimi & 36.9\% & 1.9\% & 1.0$\times$ \\
Qwen & 50.3\% & 3.5\% & 1.1$\times$ \\
\bottomrule
\end{tabular}
\caption{Inference efficiency of \ours~averaged over tasks. $N_{\text{env}}/N_{\text{total}}$: fraction of tokens from environment, processed via fast prefill rather than slow decoding. $T_{\text{exe}}/T_{\text{total}}$: fraction of total time spent on environment execution. QPM Ratio: $\text{QPM}_{\text{\ours}}/\text{QPM}_{\text{LLM}}$; values $\geq$1$\times$ indicate comparable or faster throughput.}
\label{tab:efficiency_summary}
\end{table}

\subsection{Sandbox Infrastructure}
\label{sec:sandbox_overhead}
We analyze the infrastructure overhead of these sandboxes in terms of storage, memory.
A key advantage of \ours~is its \textit{general, lightweight sandbox} design. Table~\ref{tab:sandbox_overhead} summarizes the infrastructure overhead, which is negligible in practice. For storage, typical code agent often requires task-specific environments with particular dependencies; in contrast, \ours~employs a single Docker image ($\sim$1.1 GB) shared across all tasks. Models autonomously install task-specific packages at runtime, reducing storage by orders of magnitude. For memory, each sandbox container consumes only $\sim$50 MB idle and $\sim$200 MB at peak. Even with $K=512$ concurrent sandboxes on a single DGX node, the memory overhead is only 5\% of the system RAM.

\begin{table}[h]
\centering
\small
\setlength{\tabcolsep}{2pt}
\begin{minipage}{0.48\textwidth}
\centering
\begin{tabular}{lc}
\toprule
\textbf{Dataset} & \textbf{Storage} \\
\midrule
SWE-Gym~\citep{swegym} & 6 TB \\
SWE-Smith~\citep{swesmith} & 295 GB \\
SWE-bench Verified~\citep{swe_bench} & 257 GB \\
\textbf{\ours~} & \textbf{1.1 GB} \\
\bottomrule
\end{tabular}
\end{minipage}
\hfill
\setlength{\tabcolsep}{2pt}
\begin{minipage}{0.48\textwidth}
\centering
\begin{tabular}{lcc}
\toprule
\textbf{Configuration} & \textbf{Memory} & \textbf{\% of 2TB} \\
\midrule
Per container (idle) & 50 MB & -- \\
Per container (peak) & 200 MB & -- \\
$K=64$ sandboxes & 13 GB & 0.7\% \\
$K=512$ sandboxes & 100 GB & 5\% \\
\bottomrule
\end{tabular}
\end{minipage}
\caption{\textbf{Left}: Storage overhead comparison. \ours~uses a general-purpose lightweight image, reducing storage by orders of magnitude. \textbf{Right}: Memory overhead on a DGX node with 2 TB system RAM.}
\label{tab:sandbox_overhead}
\end{table}

\section{Reinforcement Learning within Computers Enhances Generalization}
\label{sec:ise_rl}
The experiments in Section~\ref{sec:ise} demonstrate that the computer environment holds significant potential for enhancing general intelligence: strong agentic models consistently benefit across diverse domains. This raises a natural question: \textit{can we directly train LLMs within the computer environment to further unlock their potential?} Motivated by this, we propose \textbf{\ours~Reinforcement Learning (\oursrl)}, which trains models on general context-based tasks within the computer environment, enabling them to effectively explore the environment without requiring specialized agentic data.

\subsection{Method}
To train LLMs to effectively utilize computer environments for general tasks, the ideal approach should satisfy two criteria: (1) training within the computer environment to learn to explore the environment, and (2) using general-domain data to ensure broad transferability. As shown in Table~\ref{tab:rl_comparison}, existing approaches achieve some but not all of these goals. For example, vanilla RL training for LLMs on text-only tasks (hereafter LLM-RL; \citealp{rlvr}), though capable of leveraging general data, does not involve computer environment interaction; SWE-RL~\citep{swerl,deepswe2025}, which trains models on software engineering tasks within computer environments, enables environment interaction but relies on domain-specific data.

We propose training LLMs within computer environments configured with general-purpose data. Specifically, we adopt \textit{context-based tasks}: each task consists of background materials (e.g., documents) and a task objective that must be completed based on these materials. Since completing the objective depends on the provided materials, models must actively explore the computer environment to find relevant information, naturally learning to leverage its capabilities. Meanwhile, our method uses only a general-purpose computer environment without task-specific configurations, making it easy to scale. Moreover, general-purpose tasks are much simpler to curate than software engineering tasks, enabling easy data scaling~\citep{instructpt,r2e}. Overall, \oursrl~combines the benefits of computer environment-based training with general-domain data and easy scalability.

\begin{table}[h]
\centering
\small
\begin{tabular}{l|ccc}
\toprule
& \textbf{LLM-RL} & \textbf{SWE-RL} & \textbf{\oursrl} \\
\midrule
Computer Utilization & \textcolor{red}{\ding{55}} & \textcolor{green!50!black}{\ding{51}} & \textcolor{green!50!black}{\ding{51}} \\
General Domain & \textcolor{green!50!black}{\ding{51}} & \textcolor{red}{\ding{55}} & \textcolor{green!50!black}{\ding{51}} \\
Data Scalability & \textcolor{green!50!black}{\ding{51}} & \textcolor{red}{\ding{55}} & \textcolor{green!50!black}{\ding{51}} \\
Environment Scalability & N/A & \textcolor{red}{\ding{55}} & \textcolor{green!50!black}{\ding{51}} \\
\bottomrule
\end{tabular}
\caption{Comparison of different RL training paradigms. LLM-RL refers to RL training for LLMs on text-only tasks; SWE-RL refers to RL training on software engineering tasks within computer environments.}
\label{tab:rl_comparison}
\end{table}

\paragraph{Data Source.}
We source general data from context-based task datasets, specifically the seed data used for fine-tuning the synthesizer in \textbf{\textit{Instruction Pre-Training}}~\citep{instructpt}. The data covers diverse domains including encyclopedia, fiction, expert materials, academic tests, news, social media, and trivia. Each data instance has background material as context, along with a series of related tasks. The task types include free-form generation, multiple choice, and reasoning. 

\begin{figure}[!htb]
\vspace{-10pt}
\begin{center}
\includegraphics[width=\textwidth]{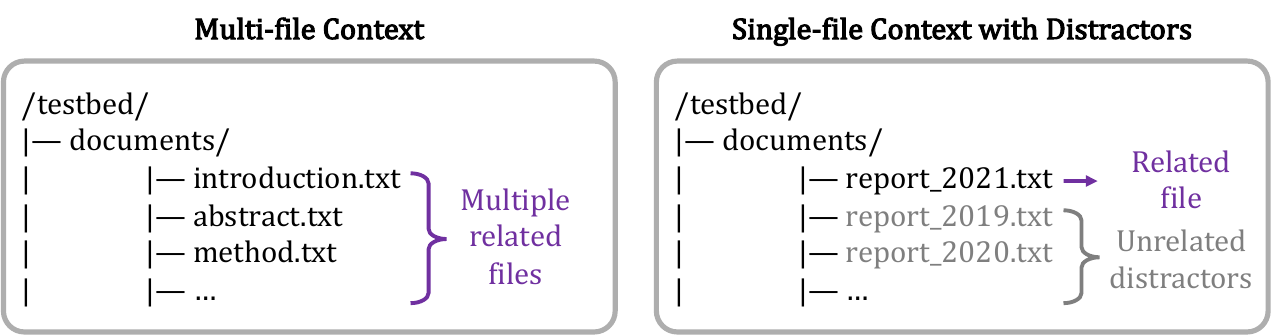}
\end{center}
\vspace{-5pt}
\caption{Environment Configuration for \oursrl: task contexts are stored as files within the sandbox. (a) Multi-document or long contexts are split into separate files. (b) Single-file contexts are supplemented with distractors.}
\label{fig:rl_context}
\end{figure}

\paragraph{Computer Environment Configuration.}
As illustrated in Figure~\ref{fig:rl_context}, we store task contexts as files within the computer environment. To enrich the context and increase the task difficulty, we employ several strategies:
\begin{itemize}[leftmargin=*,itemsep=2pt,topsep=2pt]
    \item \textbf{Multi-document or long contexts}: If a task is based on multiple documents or a single long document, we split them into separate files. For example, a research paper is divided into sections (e.g., \texttt{introduction.txt}, \texttt{methods.txt}, ...).
    \item \textbf{Single-file contexts with distractors}: If the context results in only one file, we sample multiple additional files from the same dataset as distractors, encouraging the model to navigate and filter relevant information.
\end{itemize}

\begin{wrapfigure}{r}{0.42\textwidth}
\vspace{-15pt}
\begin{center}
\includegraphics[width=0.42\textwidth]{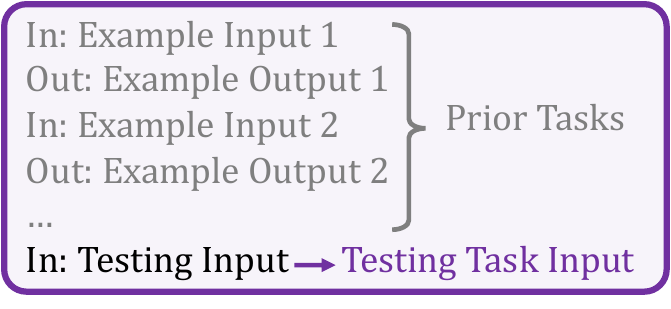}
\end{center}
\vspace{-15pt}
\caption{Task Setup. Prior related tasks are used as in-context examples.}
\label{fig:rl_task}
\vspace{-10pt}
\end{wrapfigure}

\paragraph{Task Setup.}
For each data instance, the context may correspond to multiple related tasks that depend on each other in a fixed order. As shown in Figure~\ref{fig:rl_task}, we sample one task as the testing task, using prior tasks as in-context examples in the prompt. Additionally, we inform the model in the prompt that relevant files are in \texttt{/testbed/documents/} and instruct it to write the final answer to \texttt{/testbed/answer.txt}. Upon task completion, we extract the answer from this file as the model's final output.

\paragraph{RL Training.}
Recent advances in RL for LLMs typically adopt outcome-based rewards: given a prompt, the model generates a trajectory, and the entire trajectory is rewarded based on the correctness of the final output~\citep{r1}. Our RL baseline follows this paradigm to train LLMs on context-based tasks without environmental interaction, which we refer to as LLM-RL. \oursrl~adopts the same framework, with the only difference being that trajectory generation uses \ours~mode instead of vanilla LLM mode.

\subsection{Experiments}

\paragraph{Setup}
We train from two base models that exhibit different capability levels in our evaluation in Section~\ref{sec:ise}: (1) \textit{Qwen3-4B-Instruct-2507}, a small general-purpose LLM with weaker agentic capabilities that performs worse in \ours~mode than in LLM mode; and (2) \textit{Qwen3-Coder-30B-A3B}, a code-specialized model with strong agentic abilities that already shows better performance in \ours~than in LLM mode. For benchmarks, in addition to the six non-code domains in Section~\ref{sec:ise}, we also evaluate on software engineering (SWE) tasks: SWE-bench Verified~\citep{swe_bench}, to examine whether training on general data would impair code agentic ability. Since SWE tasks inherently require a sandbox and have no LLM mode, we only report \ours~results. Evaluation details are in Appendix~\ref{app:evaluation_details}. We use rule-based functions for rewards. Detailed training settings are provided in Appendix~\ref{sec:appendix_ise_rl}.

\begin{table}[t]
\centering
\small
\setlength{\tabcolsep}{3pt}
\resizebox{\textwidth}{!}{
\begin{tabular}{p{1.4cm}ccccccccccccc}
\toprule
& \multicolumn{2}{c}{\textbf{Math}} & \multicolumn{2}{c}{\textbf{Physics}} & \multicolumn{2}{c}{\textbf{Chem.}} & \multicolumn{2}{c}{\textbf{Biomed.}} & \multicolumn{2}{c}{\textbf{Long-Cont.}} & \multicolumn{2}{c}{\textbf{Instruct.}} & \textbf{SWE} \\ \cmidrule(rl){2-3} \cmidrule(rl){4-5} \cmidrule(rl){6-7} \cmidrule(rl){8-9} \cmidrule(rl){10-11} \cmidrule(rl){12-13} \cmidrule(rl){14-14}
 & LLM & \oursbox & LLM & \oursbox & LLM & \oursbox & LLM & \oursbox & LLM & \oursbox & LLM & \oursbox & \oursbox \\
 \midrule
\multicolumn{14}{c}{\textbf{Qwen3-4B-Instruct-2507}} \\
\cmidrule(rl){1-14}
Base LLM & 46.0 & 32.5 & 22.9 & 25.6 & 56.4 & 49.3 & 10.2 & 9.4 & 8.6 & 10.5 & 33.7 & 29.0 & 11.2 \\
LLM-RL & {48.5} & 33.1 & 21.3 & 26.0 & 52.4 & 50.4 & {9.4} & 12.8 & 9.1 & 14.1 & 35.3 & 34.0 & 12.8 \\
\textit{$\Delta$} & \textcolor{green!50!black}{+2.5} & \textcolor{green!50!black}{+0.6} & \textcolor{red}{-1.6} & \textcolor{green!50!black}{+0.4} & \textcolor{red}{-4.0} & \textcolor{green!50!black}{+1.1} & \textcolor{red}{-0.8} & \textcolor{green!50!black}{+3.4} & \textcolor{green!50!black}{+0.5} & \textcolor{green!50!black}{+3.6} & \textcolor{green!50!black}{+1.7} & \textcolor{green!50!black}{+5.0} & \textcolor{green!50!black}{+1.6} \\
\rowcolor[HTML]{EAE8FD} \oursbox-RL~& {51.0} & {53.3} & {23.0} & {30.3} & {57.8} & {63.3} & 10.8 & {13.2} & {10.6} & {17.6} & {33.7} & {38.7} & {12.8} \\
\rowcolor[HTML]{EAE8FD} \textit{$\Delta$} & \textcolor{green!50!black}{+5.0} & \textcolor{green!50!black}{+20.8} & \textcolor{green!50!black}{+0.1} & \textcolor{green!50!black}{+4.7} & \textcolor{green!50!black}{+1.3} & \textcolor{green!50!black}{+14.0} & \textcolor{green!50!black}{+0.6} & \textcolor{green!50!black}{+3.8} & \textcolor{green!50!black}{+2.0} & \textcolor{green!50!black}{+7.1} & \textcolor{gray}{0.0} & \textcolor{green!50!black}{+9.7} & \textcolor{green!50!black}{+1.6} \\
\midrule
\multicolumn{14}{c}{\textbf{Qwen3-Coder-30B-A3B}} \\
\cmidrule(rl){1-14}
Base LLM & {26.0} & 41.5 & 23.9 & 28.9 & 48.7 & 54.0 & 14.4 & 18.2 & 24.6 & 23.9 & {34.7} & 40.3 & 45.0 \\
LLM-RL & 28.1 & 40.9 & 23.9 & 29.5 & 50.5 & 57.1 & {16.8} & 15.6 & {26.5} & 21.9 & 34.0 & 39.3 & 47.6 \\
\textit{$\Delta$} & \textcolor{green!50!black}{+2.1} & \textcolor{red}{-0.6} & \textcolor{gray}{0.0} & \textcolor{green!50!black}{+0.6} & \textcolor{green!50!black}{+1.8} & \textcolor{green!50!black}{+3.1} & \textcolor{green!50!black}{+2.4} & \textcolor{red}{-2.6} & \textcolor{green!50!black}{+1.9} & \textcolor{red}{-2.0} & \textcolor{red}{-0.7} & \textcolor{red}{-1.0} & \textcolor{green!50!black}{+2.6} \\
\rowcolor[HTML]{EAE8FD}\oursbox-RL~& 29.8 & {42.7} & {25.4} & {29.3} & {49.1} & {56.7} & 18.6 & {17.6} & 24.6 & {36.1} & 31.7 & {38.0} & {48.0} \\
\rowcolor[HTML]{EAE8FD} \textit{$\Delta$} & \textcolor{green!50!black}{+3.8} & \textcolor{green!50!black}{+1.2} & \textcolor{green!50!black}{+1.5} & \textcolor{green!50!black}{+0.4} & \textcolor{green!50!black}{+0.4} & \textcolor{green!50!black}{+2.7} & \textcolor{green!50!black}{+4.2} & \textcolor{red}{-0.6} & \textcolor{gray}{0.0} & \textcolor{green!50!black}{+12.2} & \textcolor{red}{-3.0} & \textcolor{red}{-2.3} & \textcolor{green!50!black}{+3.0} \\
\bottomrule
\end{tabular}
}
\vspace{-5pt}
\caption{Main results comparing \oursrl~with LLM-RL baseline, both use general context-based task data. \oursbox~denotes \ours~mode. $\Delta$ indicates performance change from Base LLM (\textcolor{green!50!black}{green} = gain, \textcolor{red}{red} = decline).}
\label{tab:ise_rl_main}
\end{table}

\paragraph{Main Results.}
Table~\ref{tab:ise_rl_main} compares \oursrl~with the LLM-RL baseline. \oursrl~demonstrates broad generalization along three axes: 
\begin{itemize}[leftmargin=*,itemsep=2pt,topsep=2pt]
    \item \textbf{Domains}: Our training uses only general context-based data, with no overlap with the training or test sets of any evaluated benchmark. Yet \oursrl~improves performance across all domains, such as Long-Context, Math, Physics, and even tasks with vastly different formats such as Instruction-Following and SWE.
    \item \textbf{Model Capabilities}: For weaker models (Qwen3-4B-Instruct), \ours~mode significantly outperforms LLM mode after \oursrl~training on most tasks (e.g., Biomed: 14.4 vs. 10.0, Instruction Following: 37.7 vs. 33.7). For stronger models (Qwen3-Coder), \oursrl~still yields consistent gains across domains.
    \item \textbf{Inference Modes}:  Surprisingly, although trained exclusively in \ours~mode, \oursrl~also improves LLM mode and even outperforms LLM-RL on most tasks, suggesting that agentic skills can transfer back to non-agentic generation. 
\end{itemize}

\paragraph{Data Source and Context Placement}
To understand the impact of training data, we compare four variants of~\oursrl~on Qwen3-4B-Instruct-2507: (1) \textbf{Math}: mathematical data from DAPO~\citep{dapo}; (2) \textbf{SWE}: software engineering data from R2E-Gym~\citep{r2e}; (3) \textbf{Gen. {in Prompt}}: our general context-based data with context placed in the prompt; and (4) \textbf{Gen. {in Sandbox}}: the same data but with context placed in the sandbox. As shown in Table~\ref{tab:ise_rl_ablation}, all variants achieve some degree of cross-domain generalization, demonstrating the broad applicability of our training paradigm. Among them, Gen. Sandbox achieves the best overall performance. Notably, the comparison between Gen. Prompt and Gen. Sandbox highlights the importance of environmental interaction: placing context in the sandbox forces the model to actively explore the environment, yielding stronger generalization than directly providing context in the prompt.

\begin{table}[h]
\centering
\small
\setlength{\tabcolsep}{2.0pt}
\resizebox{\textwidth}{!}{
\begin{tabular}{p{2.3cm}ccccccccccccc}
\toprule
 & \multicolumn{2}{c}{\textbf{Math}} & \multicolumn{2}{c}{\textbf{Physics}} & \multicolumn{2}{c}{\textbf{Chem.}} & \multicolumn{2}{c}{\textbf{Biomed.}} & \multicolumn{2}{c}{\textbf{Long-Cont.}} & \multicolumn{2}{c}{\textbf{Instruct.}} & \textbf{SWE} \\ \cmidrule(rl){2-3} \cmidrule(rl){4-5} \cmidrule(rl){6-7} \cmidrule(rl){8-9} \cmidrule(rl){10-11} \cmidrule(rl){12-13} \cmidrule(rl){14-14}
 & LLM & \oursbox & LLM & \oursbox & LLM & \oursbox & LLM & \oursbox & LLM & \oursbox & LLM & \oursbox & \oursbox \\
\midrule
Base LLM & 46.0 & 32.5 & 22.9 & 25.6 & 56.4 & 49.3 & 10.2 & 9.4 & 8.6 & 10.5 & \textbf{33.7} & 29.0 & 11.2 \\
Math & \textbf{53.5} & 51.0 & 23.4 & \textbf{31.4} & \textbf{58.2} & 60.0 & 9.6 & 13.6 & 9.5 & 11.3 & 33.0 & 37.0 & 15.2 \\
SWE & 47.5 & 36.5 & 21.3 & 21.2 & 57.8 & 47.8 & 9.4 & 9.2 & \textbf{11.6} & 9.9 & 33.0 & 33.0 & \textbf{17.4} \\
Gen. {in Prompt} & 47.7 & 37.7 & \textbf{24.0} & 26.4 & 58.0 & 58.4 & 9.6 & \textbf{16.6} & 10.2 & 15.1 & 33.0 & 34.3 & 16.2 \\
\rowcolor[HTML]{EAE8FD} Gen. {in Sandbox} & {51.0} & \textbf{53.3} & 23.0 & 30.3 & 57.8 & \textbf{63.3} & \textbf{10.8} & 13.2 & 10.6 & \textbf{17.6} & \textbf{33.7} & \textbf{38.7} & {12.8} \\
\bottomrule
\end{tabular}
}
\vspace{-5pt}
\caption{Data ablation: comparing \oursrl~with different training data (math-specific, SWE-specific, and general context-based). Gen. {in Prompt} and Gen. {in Sandbox} both use general context data but place the context differently: in prompt vs. in sandbox. \oursbox~denotes \ours~mode.}
\label{tab:ise_rl_ablation}
\end{table}

\subsection{Analysis on Generalization}
\label{sec:why_ise_improves}
To understand why \oursrl~training leads to broad generalization, we first examine how models change their computer capability usage after training, and then investigate how sandbox-mode training transfers to LLM mode. We adopt the same capability classification and quantification method as in Section~\ref{sec:analysis}.

\paragraph{Generalization across Domains.} Table~\ref{tab:ise_behavior_change} shows that after training, models exhibit increased computer capability usage across all three dimensions (external resources, file management, computation). As analyzed in Section~\ref{sec:ise}, these capabilities benefit diverse task domains, explaining why the learned exploration skills transfer broadly.

\begin{table}[h]
\centering
\small
\begin{tabular}{ll|ccc|c}
\toprule
\textbf{Model} & & \textbf{External} & \textbf{File} & \textbf{Computation} & \textbf{Avg. Turns} \\
\midrule
\multirow{2}{*}{Qwen3-4B-Instruct-2507} & Base LLM & 0.8\% & 2.9\% & 2.9\% & 23.7 \\
& \oursbox-RL & 4.1\% & 7.3\% & 7.2\% & 7.0 \\
\midrule
\multirow{2}{*}{Qwen3-Coder-30B-A3B} & Base LLM & 5.7\% & 24.1\% & 11.1\% & 9.5 \\
& \oursbox-RL & 5.7\% & {24.4}\% & 11.9\% & 10.0 \\
\bottomrule
\end{tabular}
\vspace{-5pt}
\caption{Sandbox behavior usage rate changes after \oursrl~training. Usage rate = capability invocations / total turns.}
\label{tab:ise_behavior_change}
\end{table}

\paragraph{Generalization across Model Capabilities.} As shown in Table~\ref{tab:ise_behavior_change}, weaker models (Qwen3-4B) show larger improvements: capability usage rate increases substantially while average turns decrease dramatically from 23.7 to 7.0. Recall from Section~\ref{sec:ise_analysis} that the base Qwen3-4B model ``wanders'' in the computer environment with many ineffective turns; after \oursrl~training, the model learns to accomplish tasks with fewer but more purposeful interactions. Stronger models (Qwen3-Coder) already have high capability usage, so improvements are more modest but still consistent.

\paragraph{Generalization across Inference Modes.} We analyze reasoning patterns in outputs of vanilla LLM mode before and after \oursrl~training. Specifically, we measure two categories of behaviors through pattern matching: (1) \textit{Structural Organization}: Markdown formatting elements (headers, separators, bullet points, math blocks) that indicate explicit step-by-step reasoning; and (2) \textit{Verification Behaviors}: phrases indicating self-checking (e.g., ``let's verify'', ``check that'', confirmation markers). As shown in Table~\ref{tab:llm_reasoning}, both models exhibit increased structural organization and verification behaviors after training. These reasoning patterns, learned through multi-turn environmental interaction where each action receives explicit feedback, transfer to LLM mode even without environment access.

\begin{table}[h]
\centering
\small
\begin{tabular}{lccccc}
\toprule
\multirow{2}{*}{\textbf{Model}} & \multicolumn{2}{c}{\textbf{Verification}} & \multicolumn{2}{c}{\textbf{Structure}} \\
\cmidrule(lr){2-3} \cmidrule(lr){4-5}
 & Base LLM & \oursbox-RL & Base LLM & \oursbox-RL \\
\midrule
Qwen3-4B-Instruct-2507 & 20.22 & 36.91 & 19.13 & 20.64 \\
Qwen3-Coder-30B-A3B & 0.77 & 0.88 & 10.30 & 16.12 \\
\bottomrule
\end{tabular}
\vspace{-5pt}
\caption{Reasoning pattern changes in outputs of vanilla LLM mode after \oursrl~training. Values are counts per response.}
\label{tab:llm_reasoning}
\end{table}

\section{LLMs within Computer Environments Go Beyond Text Generation}
\label{sec:beyond}

Previous sections evaluate \ours~on tasks where both vanilla LLMs and \ours~can produce outputs for end-to-end comparison. However, \ours~also enables capabilities that are \textbf{\textit{fundamentally impossible}} for standalone LLMs. By granting LLMs access to the computer environment, \ours~transcends the text-in-text-out paradigm and unlocks new possibilities:

\begin{itemize}[leftmargin=*,itemsep=2pt]
    \item \textbf{Cross-Modal Capabilities}: LLMs are confined to text-in-text-out, but \ours~enables processing and generating images, videos, audio, and interactive applications by orchestrating specialized software within the sandbox.
    \item \textbf{File-Level Operations}: Rather than describing what a file should contain, \ours~directly produces actual files (e.g., \texttt{.png}, \texttt{.mp4}, \texttt{.wav}, \texttt{.html})that users can immediately use, with grounded feedback from real execution.
    \item \textbf{Autonomous Tool Acquisition}: Unlike predefined tool-use where LLMs call fixed APIs, \ours~enables LLMs to autonomously discover, install, and learn to use arbitrary software libraries on demand, effectively granting unlimited tool access.
\end{itemize}

\begin{figure}[!htb]
    \centering
    \includegraphics[width=\linewidth]{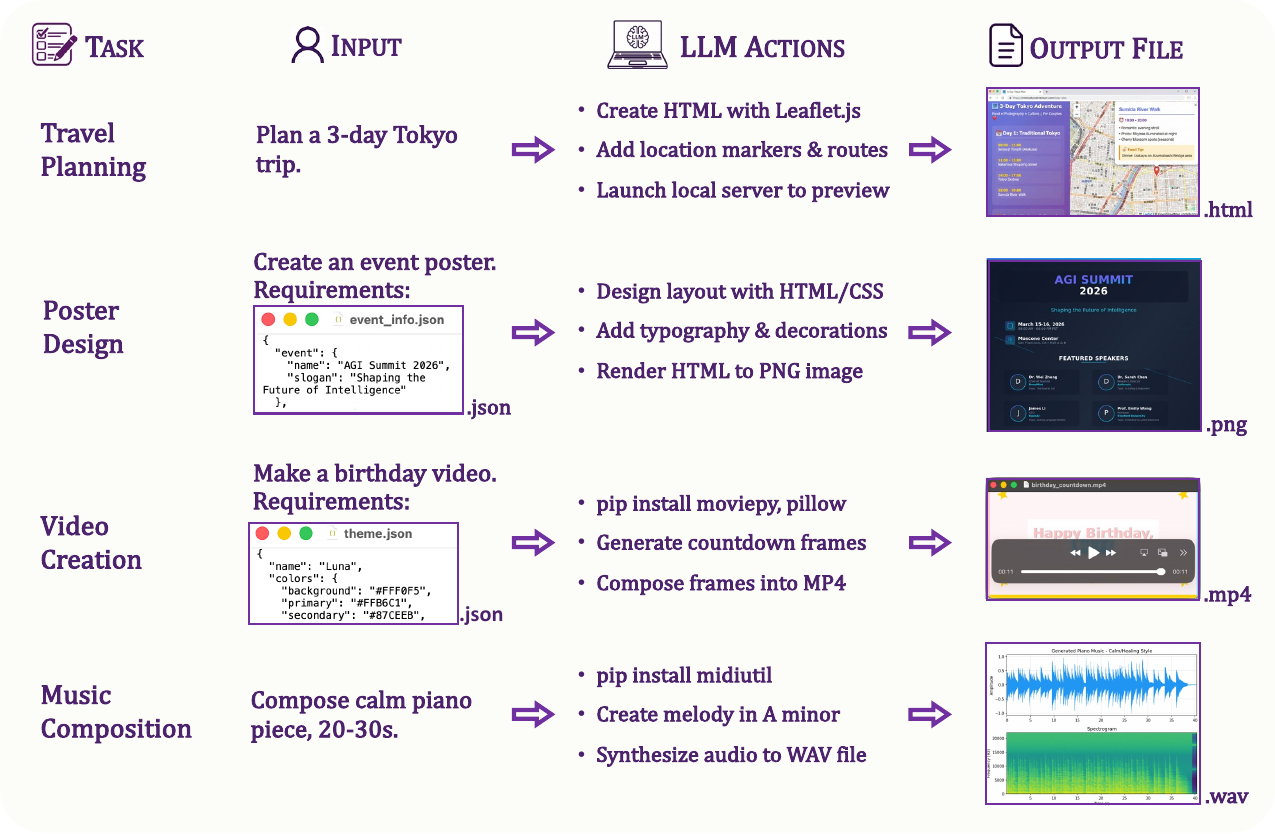}
    \vspace{-20pt}
    \caption{\ours~transcends the text-in-text-out paradigm. By granting LLMs access to a basic virtual computer, they can autonomously install tools, write and execute programs, and produce usable files: interactive webpages (\texttt{.html}), images (\texttt{.png}), videos (\texttt{.mp4}), and audio (\texttt{.wav}).}
    \label{fig:cross_modal}
\end{figure}

\paragraph{Case Studies.}
We demonstrate these capabilities through four representative examples in Figure~\ref{fig:cross_modal}. Full trajectories and interactive demos are available at our project page.

\textbf{Case 1: Travel Planning → Interactive Map.}
Given a natural language query for a 3-day Tokyo trip itinerary, the agent installs Leaflet.js, designs a data structure for 12 locations, generates JavaScript for markers with popups and color-coded route polylines, producing a fully functional \texttt{map.html} with clickable markers and day-by-day visualization.

\textbf{Case 2: Event Specification → Conference Poster.}
From a JSON file containing event details (``AGI Summit 2026'', venue, speakers, sessions), the agent designs an SVG layout with gradient backgrounds, implements typography hierarchy, and converts to PNG via CairoSVG~\footnote{https://github.com/Kozea/CairoSVG}, outputting professional \texttt{poster.svg} and \texttt{poster.png} files.

\textbf{Case 3: Theme Configuration → Animated Video.}
Given a JSON theme specifying recipient name, color palette, and aesthetic style, the agent generates 360 frames using PIL with animated decorations, compiles them at 30fps via moviepy~\footnote{https://github.com/Zulko/moviepy}, producing a 11-second \texttt{birthday\_countdown.mp4} .

\textbf{Case 4: Style Description → Original Music.}
From a natural language request for a calm piano piece in A minor, the agent uses midiutil~\footnote{https://github.com/MarkCWirt/MIDIUtil} to compose melody and chord progressions, renders audio via FluidSynth~\footnote{https://github.com/FluidSynth/fluidsynth}, outputting \texttt{composition.mid}, \texttt{preview.wav}, and \texttt{sheet\_music.md}.

\paragraph{Discussion.}
While these examples demonstrate the potential of \ours~to achieve general intelligence beyond text generation, we acknowledge that current results have limitations. The generated videos are limited to simple 11-second animations without complex scenes. The composed music, though structurally correct, lacks the expressiveness and creativity of human compositions. The posters follow basic design principles but may not match professional graphic design quality.

Nevertheless, these cases reveal a promising direction: as LLMs become more capable and computer environments more sophisticated, \ours~could evolve into a truly general-purpose digital creation system. We believe this paradigm, LLMs interacting with computational environments rather than generating text in isolation, represents a compelling path toward general intelligence.

\section{Conclusion and Future Work}

\paragraph{Computer Environments as Default Inference Infrastructure}
We introduce \ours, a paradigm that grants LLMs access to a virtual computer and show that strong LLMs exhibit emergent capabilities to leverage this environment for general tasks. Looking forward, we envision \ours~becoming the default paradigm for serving LLMs: analytical tasks gain verifiable computation, long-context tasks benefit from file-based management, and creative tasks yield actual outputs (e.g., images, videos and applications) rather than text descriptions. We anticipate computer environments will become standard infrastructure, transforming LLMs from text generators into general-purpose digital workers.

\paragraph{Computer Environments as an Agentic Capability Benchmark}
Beyond serving as an inference paradigm, \ours~naturally provides a standardized testbed for evaluating agentic capabilities. Unlike existing benchmarks that focus on specific downstream tasks~\citep{swe_bench,tau_bench,browsecomp}, \ours~measures fundamental skills like exploration, tool use, and self-verification through a unified framework. The metric $\Delta = \text{\ours} - \text{LLM}$ offers a meaningful indicator: it quantifies how effectively a model can leverage computational environments, revealing agentic potential that raw LLM performance alone cannot capture.

\paragraph{Computer-Native Model Training}
We propose \oursrl, a lightweight RL method that trains computer interaction as a transferable skill using only general, non-agentic data. Looking ahead, we advocate for \textit{computer-native} models where computer interaction becomes a first-class training objective, not only through large-scale RL with real environmental feedback, but also by incorporating computer-style reasoning into the pretraining stage itself.

\section*{Acknowledgments}
The first author would like to thank Yejie Wang, Lisheng Huang, and Shuang Sun for helpful discussions, and the R2E-Gym~\citep{r2e}, DeepSWE~\citep{deepswe2025}, and rLLM~\citep{rllm} teams for their valuable open-source contributions.

\bibliography{colm2026_conference}
\bibliographystyle{colm2026_conference}

\appendix

\section{Sandbox Implementation}
\label{sec:sandbox_implementation}
 We build upon the sandbox framework of R2E-Gym~\citep{r2e}, adapting it for general-purpose exploration across diverse non-code domains. Complete specifications of the tools available within the sandbox are detailed below.

\begin{tcolorbox}[
    title=bash,
    colback=white,
    colframe=lavender,
    coltitle=gray!40!black,
    fonttitle=\bfseries\ttfamily,
    arc=1mm,
    boxrule=0.6mm,
    left=1mm,
    right=1mm,
    top=1mm,
    bottom=1mm,
]
\small
\textbf{Description}: Execute a bash command in the terminal within a persistent shell session.
\begin{itemize}[leftmargin=*]
\itemsep0em
    \item \textit{One command at a time}: You can only execute one bash command at a time. If you need to run multiple commands sequentially, use \texttt{\&\&} or \texttt{;} to chain them together.
    \item \textit{Persistent session}: Commands execute in a persistent shell session where environment variables, virtual environments, and working directory persist between commands.
    \item \textit{Soft timeout}: Commands have a soft timeout of 10 seconds, once that's reached, you have the option to continue or interrupt the command.
    \item \textit{Output truncation}: If the output exceeds a maximum length, it will be truncated before being returned.
\end{itemize}

\textbf{Parameters}:
\begin{itemize}[leftmargin=*]
\itemsep0em
    \item \texttt{command} (string, required): The bash command to execute. For example: \texttt{python my\_script.py}.
\end{itemize}
\end{tcolorbox}

\begin{tcolorbox}[
    title=file\_editor,
    colback=white,
    colframe=lavender,
    coltitle=gray!40!black,
    fonttitle=\bfseries\ttfamily,
    arc=1mm,
    boxrule=0.6mm,
    left=1mm,
    right=1mm,
    top=1mm,
    bottom=1mm,
]
\small
\textbf{Description}: Custom editing tool for viewing, creating and editing files.
\begin{itemize}[leftmargin=*]
\itemsep0em
    \item State is persistent across command calls and discussions with the user.
    \item If \texttt{path} is a file, \texttt{view} displays the result of applying \texttt{cat -n}. If \texttt{path} is a directory, \texttt{view} lists non-hidden files and directories up to 2 levels deep.
    \item The \texttt{create} command cannot be used if the specified \texttt{path} already exists as a file.
    \item For the \texttt{str\_replace} command, the \texttt{old\_str} parameter should match EXACTLY one or more consecutive lines from the original file.
\end{itemize}

\textbf{Parameters}:
\begin{itemize}[leftmargin=*]
\itemsep0em
    \item \texttt{command} (string, required): The command to run. Allowed options are: \texttt{view}, \texttt{create}, \texttt{str\_replace}, \texttt{insert}.
    \item \texttt{path} (string, required): Absolute path to file or directory.
    \item \texttt{file\_text} (string, optional): Required for \texttt{create} command.
    \item \texttt{old\_str} (string, optional): Required for \texttt{str\_replace} command.
    \item \texttt{new\_str} (string, optional): The replacement string for \texttt{str\_replace}, or the string to insert for \texttt{insert}.
    \item \texttt{insert\_line} (integer, optional): Required for \texttt{insert} command.
    \item \texttt{view\_range} (array, optional): Line range for \texttt{view} command, e.g., \texttt{[11, 12]}.
\end{itemize}
\end{tcolorbox}

\begin{tcolorbox}[
    title=finish,
    colback=white,
    colframe=lavender,
    coltitle=gray!40!black,
    fonttitle=\bfseries\ttfamily,
    arc=1mm,
    boxrule=0.6mm,
    left=1mm,
    right=1mm,
    top=1mm,
    bottom=1mm,
]
\small
\textbf{Description}: Finish the interaction when the task is complete OR if the assistant cannot proceed further with the task.

\textbf{Parameters}: No parameters are required for this function.
\end{tcolorbox}

\section{Model Configurations}
\label{app:model_configuration}
Table~\ref{tab:model_configs} summarizes the inference configurations used for each model. The maximum turn is set as 100. The maximum generation length per turn is set to 65,536 tokens, except for Claude-Sonnet-4.5-Think which is limited to 64,000 tokens due to API constraints. For vanilla LLM mode, this represents the maximum tokens generated in a single response given the prompt. For \ours, this limit applies to each turn, and the total trajectory length (including prompt, model output and environment output) is also capped at the same value, except for the long-context understanding task where we set the trajectory limit to 131,072 tokens to accommodate the longer context.

\begin{table}[h]
\begin{center}
\small
\resizebox{\textwidth}{!}{
\begin{tabular}{lcccccc}
\toprule
\textbf{Model} & \textbf{Temperature} & \textbf{Top\_p} & \textbf{Min\_p} & \textbf{Top\_k} & \textbf{Rep. Penalty} & \textbf{Backend}\\
\midrule
Claude-Sonnet-4.5-Think & 1.0 & - & - & - & - & API \\
GPT-5 & 1.0 & - & - & - & - & API \\
DeepSeek-V3.2-Thinking & 1.0 & 0.95 & - & - & - & SGLang \\
MiniMax-M2 & 1.0 & 0.95 & - & 40 & - & vLLM \\
Kimi-K2-Thinking & 1.0 & - & - & - & - & SGLang \\
Qwen3-Coder-30B-A3B & 0.7 & 0.80 & 0.0 & 20 & 1.05 & vLLM \\
Qwen3-4B-Instruct-2507 & 0.7 & 0.80 & 0.0 & 20 & - & vLLM \\
\bottomrule
\end{tabular}}
\caption{Inference configurations for each model. We use the recommended sampling parameters by each model supplier. ``-'' indicates the parameter is not applicable or uses the default value. The thinking budget for Claude is 60,000 tokens.}
\label{tab:model_configs}
\end{center}
\end{table}

\section{Evaluation Details}
\label{app:evaluation_details}
We evaluate on six non-code domains and one code domain, summarized in Table~\ref{tab:benchmark_summary}. The system prompt largely follows the one in Appendix~\ref{sec:appendix_prompt}, with minor domain-specific adjustments and final answer formatting instructions. Please refer to our released code for the exact prompts.

\begin{table}[h]
\begin{center}
\small
\resizebox{\textwidth}{!}{
\begin{tabular}{llcc}
\toprule
\textbf{Domain} & \textbf{Benchmark} & \textbf{\# Problems} & \textbf{Evaluation} \\
\midrule
Mathematics & AIME25~\citep{AIME} & 30 $\times$ 16 & Math-Verify \\
Physics & UGPhysics~\citep{ugphysics} & 650 & LLM Judge \\
Chemistry & ChemBench~\citep{chembench} & 450 & Exact Match \\
Biomedicine & MedXpertQA~\citep{medxpertqa} & 500 & Exact Match \\
Long-Context & AA-LCR~\citep{aalcr} & 100 $\times$ 4 & LLM Equality Checker \\
Instruction Following & IFBench~\citep{ifbench} & 300 & Rule-based (Loose) \\
Software Engineering & SWE-bench Verified~\citep{swe_bench} & 500 & Rule-based \\
\bottomrule
\end{tabular}
}
\caption{Summary of evaluation benchmarks. ``$\times$ N'' indicates each problem is repeated N times.}
\label{tab:benchmark_summary}
\end{center}
\end{table}

\paragraph{Mathematics.}
We use all 30 problems from the 2025 American Invitational Mathematics Examination (AIME25), which tests olympiad-level mathematical reasoning. Given the small dataset size, we repeat each problem 16 times and report average accuracy. The prompt includes ``Please reason step by step, and put your final answer within \texttt{\textbackslash boxed\{\}}.'' and we use Math-Verify~\citep{mathverify} for evaluation.

\paragraph{Physics.}
UGPhysics is a comprehensive benchmark for evaluating physics problem-solving at the undergraduate level, spanning 13 core subjects. We sample 50 problems from each subject, yielding 650 problems in total. Responses are evaluated using an LLM-based judge~\citep{llmjudge} with Qwen3-235B-A22B-Instruct-2507~\citep{qwen3}.

\paragraph{Chemistry.}
ChemBench assesses chemistry competency through single-choice questions across nine core tasks. We sample 50 problems from each sub-domain (450 total) and use exact match for evaluation.

\paragraph{Biomedicine.}
MedXpertQA is designed to evaluate expert-level medical knowledge and reasoning through multiple-choice questions. We use only the text-based questions, sampling 500 instances, and evaluate via exact match.

\paragraph{Long-Context Understanding.}
AA-LCR contains 100 challenging questions requiring multi-document reasoning, with each document set averaging approximately 100K tokens. Answers must be derived through reasoning rather than direct retrieval. In \ours~mode, each problem is initialized with its own sandbox environment, where all related documents are stored as text files in \texttt{/testbed/documents/}, each named after its original title. We repeat each problem 4 times and report average accuracy. Following the original work, we use an LLM-based equality checker for evaluation with Qwen3-235B-A22B-Instruct-2507~\citep{qwen3}.

\paragraph{Instruction Following.}
IFBench tests precise instruction-following across 58 diverse, verifiable constraints. We use the single-turn subset (300 questions) and evaluate with the official code in loose mode, which handles formatting variations by checking multiple output forms.

\paragraph{Software Engineering.}
SWE-bench is a comprehensive benchmark for software engineering tasks, including code generation, debugging, and comprehension. We use the verified subset (500 problems) and evaluate using the official rule-based evaluation script. We leverage the SWE-bench sandbox setup from R2E-Gym~\citep{r2e} for code execution. The system prompt follows OpenHands~\citep{openhands}, and the toolset is the same as described in Section~\ref{sec:workflow}.

\section{Computer Capability Classification}
\label{app:capability_classification}
We classify model actions into three capability categories through pattern matching on bash commands and Python code. Table~\ref{tab:behavior_patterns} summarizes the classification patterns for each category.

\begin{table}[h]
\begin{center}
\small
\begin{tabular}{l|l|l}
\toprule
\textbf{Category} & \textbf{Pattern Type} & \textbf{Examples} \\
\midrule
\multirow{4}{*}{External Resources} & Package installation & \texttt{pip install}, \texttt{apt-get install} \\
& HTTP requests & \texttt{requests.get}, \texttt{curl}, \texttt{wget} \\
& Web scraping & \texttt{BeautifulSoup}, \texttt{selenium} \\
& Domain libraries & \texttt{rdkit}, \texttt{biopython}, \texttt{pubchempy} \\
\midrule
\multirow{4}{*}{File Management} & Python file I/O & \texttt{open()}, \texttt{json.load}, \texttt{pd.read\_csv} \\
& Shell file commands & \texttt{cat}, \texttt{grep}, \texttt{find}, \texttt{head/tail} \\
& Path operations & \texttt{os.path}, \texttt{pathlib}, \texttt{glob} \\
& Data serialization & \texttt{pickle.load}, \texttt{np.load/save} \\
\midrule
\multirow{4}{*}{Computation} & Numerical solvers & \texttt{scipy.optimize}, \texttt{fsolve}, \texttt{minimize} \\
& Integration & \texttt{odeint}, \texttt{solve\_ivp}, \texttt{quad} \\
& Iterative algorithms & Large loops (\texttt{range(N)} where $N>100$), \texttt{while} loops \\
& Combinatorics & \texttt{itertools.permutations/combinations} \\
\bottomrule
\end{tabular}
\caption{Pattern matching rules for detecting computer capability usage.}
\label{tab:behavior_patterns}
\end{center}
\end{table}

For each trajectory, we extract all code blocks from model actions (both Python scripts and bash commands) and apply these patterns. The capability usage rate is computed as the number of turns containing at least one matched pattern divided by the total number of interaction turns.

\section{\oursrl~Training Details}
\label{sec:appendix_ise_rl}
We train \oursrl~following the training framework of DeepSWE~\citep{deepswe2025} in rLLM~\citep{rllm}. Based on this, we penalize excessively long trajectories: if the model exceeds maximum turns/tokens without submitting an answer, the episode is terminated with zero reward. 
The sandbox configuration is identical to that described in Section~\ref{sec:ise}. Table~\ref{tab:ise_rl_hyperparams} summarizes the key hyperparameters. 

\paragraph{Reward Design.}
We use rule-based reward functions tailored to each task type: (1) for multiple-choice tasks, we assign a positive reward for selecting the correct option and 0 otherwise; when multiple correct options exist, we use F1 score as the reward; (2) for free-form generation tasks, we use ROUGE-L~\citep{rouge} score; (3) for tasks with binary correctness (e.g., math problems), we use a simple binary reward (+1 for correct, 0 for incorrect). 

\begin{table}[h]
\begin{center}
\small
\begin{tabular}{lcc}
\toprule
\textbf{Hyperparameter} & \textbf{Qwen3-4B-Instruct-2507} & \textbf{Qwen3-Coder-30B-A3B} \\
\midrule
RL Algorithm & GRPO++ & GRPO++ \\
Learning Rate & 1e-6 & 1e-6 \\
Train (Prompt) Batch Size & 8 & 8 \\
Update mini batch size & 8 & 8 \\
Rollouts per Prompt & 8 & 8 \\
Train Steps & 150 & 50 \\
Max Turns & 100 & 100 \\
KL Reward/Loss & None & None \\
Rollout Temperature & 1.0 & 1.0 \\
Rollout Top\_p & 0.8 & 0.8 \\
Rollout Top\_k & 20 & 20 \\
Max Response Length & 65,536 Tokens & 65,536 Tokens \\
\bottomrule
\end{tabular}
\caption{Hyperparameters for \oursrl~training. GRPO++ refers to the variants of GRPO~\citep{deepseekmath} used in DeepSWE~\citep{deepswe2025}. Update mini batch size indicates the batch size used for each policy update step, we set it as the same as the train batch size, meaning we perform one update per batch (i.e., on-policy).}
\label{tab:ise_rl_hyperparams}
\end{center}
\end{table}

\section{Prompt for~\ours}
\label{sec:appendix_prompt}
The prompts used in our experiments are shown in Figure~\ref{fig:instance_prompt} and Figure~\ref{fig:system_prompt}. This represents a minimal baseline prompt that establishes the core environment interaction protocol. The prompt can be easily adapted to different use cases by modifying the input/output format specifications (e.g., changing the output file path or format), adding domain-specific instructions, or incorporating additional tools.

\begin{figure}[h]
\begin{tcolorbox}[
    title=Instance Prompt Template for~\ours,
    colback=white,
    colframe=lavender,
    coltitle=gray!40!black,
    fonttitle=\bfseries,
    arc=1mm,
    boxrule=0.6mm,
    left=1mm,
    right=1mm,
    top=1mm,
    bottom=1mm,
]
\small
\ttfamily
\textbf{<problem>}\\
\{problem\_statement\}\\
\textbf{</problem>}\\
\\
Please solve this problem.\\
\\
\textbf{<OUTPUT\_INSTRUCTIONS>}\\
- If the task requires a specific answer (e.g., a number, text, or computation result): write the final answer to /testbed/output/answer.txt (plain text, answer only, no explanations)\\
- If the task requires creating a project, code, or multiple files: save all files directly to /testbed/output/\\
\textbf{</OUTPUT\_INSTRUCTIONS>}\\
\\
Working directory: /testbed\\
Input files (if any): /testbed/input\\
Output directory: /testbed/output
\end{tcolorbox}
\caption{The Instance Prompt Template. The \texttt{\{problem\_statement\}} placeholder is filled with the actual problem description for each task instance.}
\label{fig:instance_prompt}
\end{figure}

\begin{figure}[h]
\begin{tcolorbox}[
    title=System Prompt for~\ours,
    colback=white,
    colframe=lavender,
    coltitle=gray!40!black,
    fonttitle=\bfseries,
    arc=1mm,
    boxrule=0.6mm,
    left=1mm,
    right=1mm,
    top=1mm,
    bottom=1mm,
]
\small
\ttfamily
You are an expert specializing in solving complex problems using code.\\
\\
\textbf{<TASK>}\\
You need to complete the given task by following the instructions precisely.\\
\textbf{</TASK>}\\
\\
\textbf{<DIRECTORIES>}\\
- Working directory: /testbed\\
- Input directory: /testbed/input~~<-{}- User-provided input files/assets are here\\
- Output directory: /testbed/output~~<-{}- Put ALL your outputs here\\
\textbf{</DIRECTORIES>}\\
\\
\textbf{<WORKFLOW>}\\
1. Read the problem carefully\\
2. Check /testbed/input for any input files if the task mentions them\\
3. Analyze the problem and determine the solution approach\\
4. \textbf{MUST write code to a file and execute it} - DO NOT just think about the answer, use print statements directly, or hardcode answers\\
5. Save all outputs to /testbed/output directory\\
\textbf{</WORKFLOW>}\\
\\
\textbf{<IMPORTANT\_NOTES>}\\
- Use \texttt{bash} to run scripts or commands\\
- Use \texttt{file\_editor} to view, create and edit files\\
- Use \texttt{finish} to finish once you have completed the task\\
\textbf{</IMPORTANT\_NOTES>}\\
\\
\textbf{<ENCOURAGED\_APPROACHES>}\\
- You have full access to this isolated environment - feel free to install packages, create files, run experiments, etc.\\
- Explore diverse problem-solving approaches: use libraries, tools, external data, computations, simulations, or any method that helps\\
- The environment is sandboxed for your use - be creative and try different computational strategies\\
- The more comprehensive and computational your approach, the better\\
\textbf{</ENCOURAGED\_APPROACHES>}\\
\\
\textbf{<ANTI\_HARDCODING>}\\
STRICTLY PROHIBITED: Do not use large comment blocks or print statements for natural language thinking (e.g., `\# Let me think step by step...' or `print("First, I need to consider...")'). Do not hardcode answers like `answer = "A"' or `return 'A''. You must derive the final result through actual computational logic, mathematical operations, and programmatic analysis of the problem data.\\
\textbf{</ANTI\_HARDCODING>}
\end{tcolorbox}
\caption{The System Prompt used for~\ours.}
\label{fig:system_prompt}
\end{figure}

\end{document}